%% file: main.tex
\def\year{2023}\relax
\documentclass[letterpaper]{article} 
\usepackage{aaai23}  
\usepackage{times}  
\usepackage{helvet}  
\usepackage{courier}  
\usepackage[hyphens]{url}  
\usepackage{graphicx} 
\usepackage{natbib}  
\usepackage{caption}

\usepackage[utf8]{inputenc}
\usepackage{booktabs}
\usepackage{amsfonts}
\usepackage{nicefrac}
\usepackage{microtype}
\usepackage[table]{xcolor}
\usepackage{mathtools,bbm}
\usepackage{color}
\usepackage{multirow}
\usepackage{subcaption}
\usepackage{makecell}
\usepackage{cleveref}
\usepackage{textcomp}
\usepackage{gensymb}
\usepackage{marginnote}
\usepackage{enumitem}

\definecolor{Gray}{gray}{0.9}

\newcommand{\ie}{\textit{i}.\textit{e}.}
\newcommand{\eg}{\textit{e}.\textit{g}.}

\newcommand{\reviewer}[3]{
	\expandafter\newcommand\csname #1\endcsname[1]{
		\textcolor{#3}{[#2: ##1]}
	}
}

\DeclareCaptionStyle{ruled}{labelfont=normalfont,labelsep=colon,strut=off} 
\usepackage{algorithm}
\usepackage{algorithmic}

\usepackage{newfloat}
\usepackage{listings}
\floatstyle{ruled}
\newfloat{listing}{tb}{lst}{}
\floatname{listing}{Listing}

\title{Improving Robust Fairness via Balance Adversarial Training}

\author{
    Chunyu Sun\textsuperscript{$1$}, Chenye Xu\textsuperscript{$1$}, Chengyuan Yao\textsuperscript{$1$}, Siyuan Liang\textsuperscript{$2$}, Yichao Wu\textsuperscript{$1$}, Ding Liang\textsuperscript{$1$},\\
    Xianglong Liu\textsuperscript{$3$},
    Aishan Liu \textsuperscript{$3$}
}
\affiliations{
    \textsuperscript{$1$}SenseTime Research\\
    \textsuperscript{$2$}Institute of Information Engineering, Chinese Academy of Sciences, Beijing, China\\
    \textsuperscript{$3$}Beihang University, Beijing, China\\
    \texttt{\{sunchunyu,xuchenye,yaochengyuan,wuyichao,liangding\}@sensetime.com, liangsiyuan@iie.ac.cn, \{xlliu, liuaishan\}@buaa.edu.cn}
}

\begin{document}

\maketitle

\input{texts/0_abstract}
\input{texts/1_introduction}

\input{texts/2_relatedwork}
\input{texts/3_method}
\input{texts/4_experiments}

\input{texts/5_analysis}
\input{texts/5_conclusion}

\bibliography{reference}

\end{document}

%% file: texts/0_abstract.tex
\begin{abstract}

Adversarial training (AT) methods are effective against adversarial attacks, yet they introduce severe disparity of accuracy and robustness between different classes, known as the \emph{robust fairness problem}. Previously proposed Fair Robust Learning (FRL) adaptively reweights different classes to improve fairness. However, the performance of the better-performed classes decreases, leading to a strong performance drop. In this paper, we observed two unfair phenomena during adversarial training: different difficulties in generating adversarial examples from each class (source-class fairness) and disparate target class tendencies when generating adversarial examples (target-class fairness). From the observations, we propose Balance Adversarial Training (BAT) to address the robust fairness problem. Regarding source-class fairness, we adjust the attack strength and difficulties of each class to generate samples near the decision boundary for easier and fairer model learning; considering target-class fairness, by introducing a uniform distribution constraint, we encourage the adversarial example generation process for each class with a fair tendency. Extensive experiments conducted on multiple datasets (CIFAR-10, CIFAR-100, and ImageNette) demonstrate that our method can significantly outperform other baselines in mitigating the robust fairness problem (+5-10\% on the worst class accuracy)\footnote{Our code 
will be available upon paper publication}.

\end{abstract}

%% file: texts/1_introduction.tex
\section{Introduction}

Deep neural networks (DNNs) are vulnerable to adversarial attacks \cite{szegedy2014intriguing,DBLP:journals/corr/GoodfellowSS14} which fool model predictions by adding imperceptible perturbations to natural examples. 
To defend against adversarial attacks, many defense techniques are designed \cite{DBLP:conf/cvpr/XieWMYH19,cohen2019certified,jeong2020consistency}. In particular, adversarial training \cite{madry2018towards, zhang19p} that injects adversarial examples during training has been proved to be the most effective methods against adversarial attacks. 

However, adversarial training suffers from the \emph{robust fairness problem}, where the adversarially trained models make a severe disparity in accuracy and robustness among different classes \cite{DBLP:conf/icml/XuLLJT21}. For example, an adversarially trained ResNet-18 model on CIFAR-10 has significantly lower clean and robust accuracy on class \texttt{cat} than other classes; in contrast, each class has a similar accuracy during the standard training (see Figure \ref{fig:concept-fig}). This phenomenon is firstly defined by \cite{DBLP:conf/icml/XuLLJT21} and further theoretically justified by studying a binary classification task under a Gaussian mixture distribution. To mitigate the robust fairness problem, they proposed Fair-Robust-Learning (FRL), which adaptively re-weights each class during training to balance the performance of each class. However, at a closer inspection, we found that this robust fairness is achieved by reducing the performance of other previously better performed classes, leading to a reduction in both clean and robust accuracy (Figure \ref{fig:concept-fig}).

 \input{resources/fig_1_concept_figure}

In this paper, we conjecture that the mechanisms of the adversarial example generation process during adversarial training are related to the robust fairness problem, which cannot be mitigated by a class re-weighting scheme. More specifically, we found two key observations that are fundamental to the robust fairness during AT as (1) \emph{source-class fairness}: samples from different classes have different difficulties and require different perturbation budgets for adversarial example generation; (2) \emph{target-class fairness}: the targets of the generated adversarial examples are biased and yield a clear tendency towards specified classes. 
Motivated by the above observation, we propose the Balance Adversarial Training (BAT) framework to mitigate the robust fairness problem by simultaneously addressing source-class and target-class fairness issues. To mitigate source-class fairness, we balance the strength of adversarial attacks on each class with adaptive perturbations so that we could bring samples to decision boundaries which would be easier and fairer for models to learn; to balance target-class fairness, we force the generated adversarial examples to follow a uniform distribution towards target classes, so that we could yield a fairer classifier not influenced by the tendency of adversarial targets. Extensive experiments have been conducted on CIFAR-10, CIFAR-100, and ImageNette, demonstrating that BAT improves robust fairness while preserving both accuracy and robustness. In particular, our method outperforms other baselines by large margins and improves the worst class error rate of 6.31\% on average. 
{Our \textbf{contributions} can be summarized as:
\begin{itemize}
    \item We discover the source-class and target-class fairness phenomena as the related cause of the robust fairness problem for AT.
    \item Based on our observation, we propose a novel AT framework named BAT to mitigate the fairness problem, where we balance the source-class and target-class fairness.
    \item Extensive experiments on several datasets have been conducted, which demonstrate the superiority of our approach compared to other baselines.
\end{itemize}
}

%% file: resources/fig_1_concept_figure.tex
\begin{figure}[t]
\centering
\begin{subfigure}[t]{0.49\linewidth}
		\includegraphics[width=\textwidth,trim=0 0 0 0,clip]{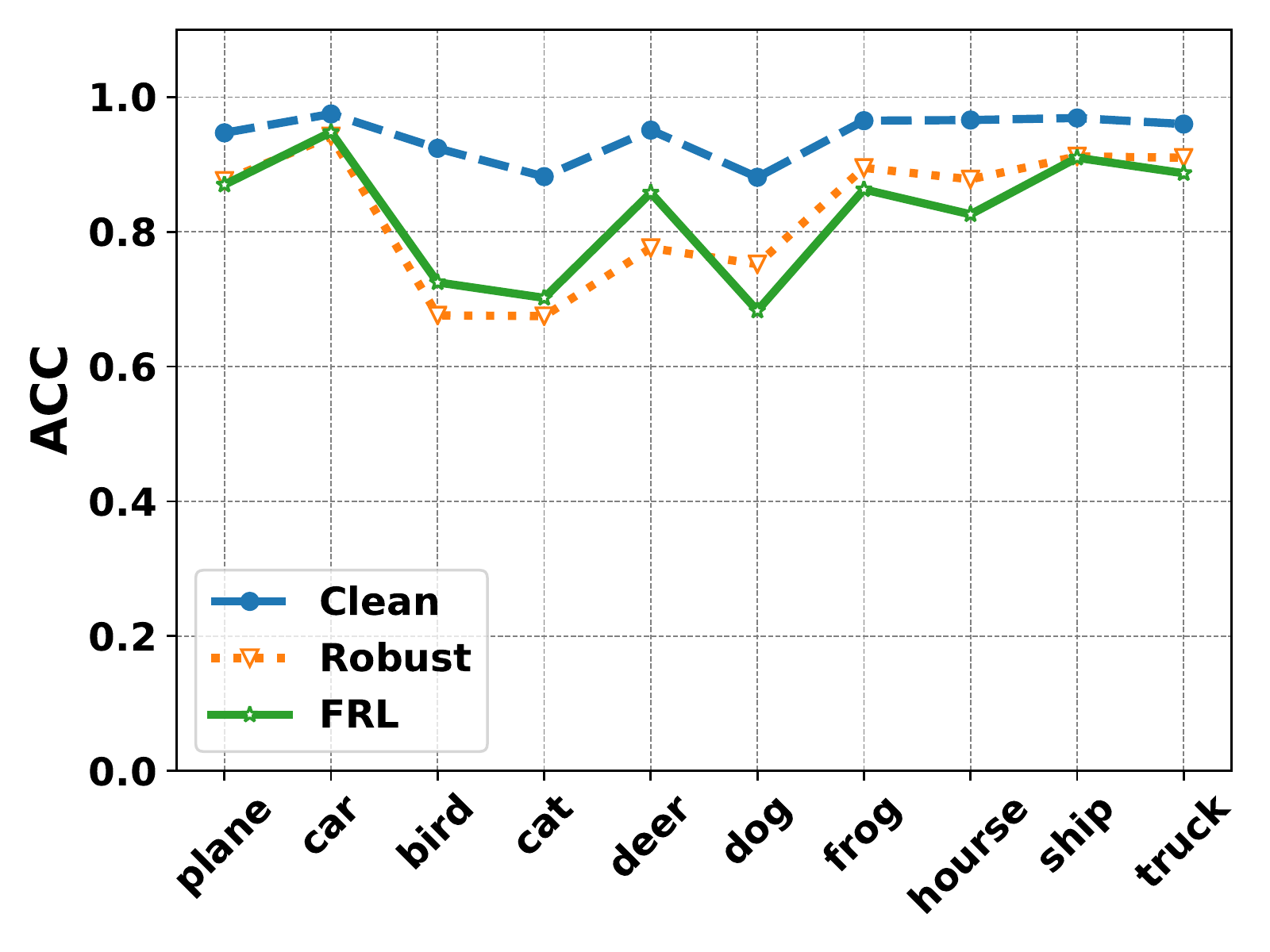}
		\caption{Clean accuracy}
		\label{fig:robust_fairness}
\end{subfigure}
\begin{subfigure}[t]{0.49\linewidth}
		\includegraphics[width=\textwidth,trim=0 0 0 0,clip]{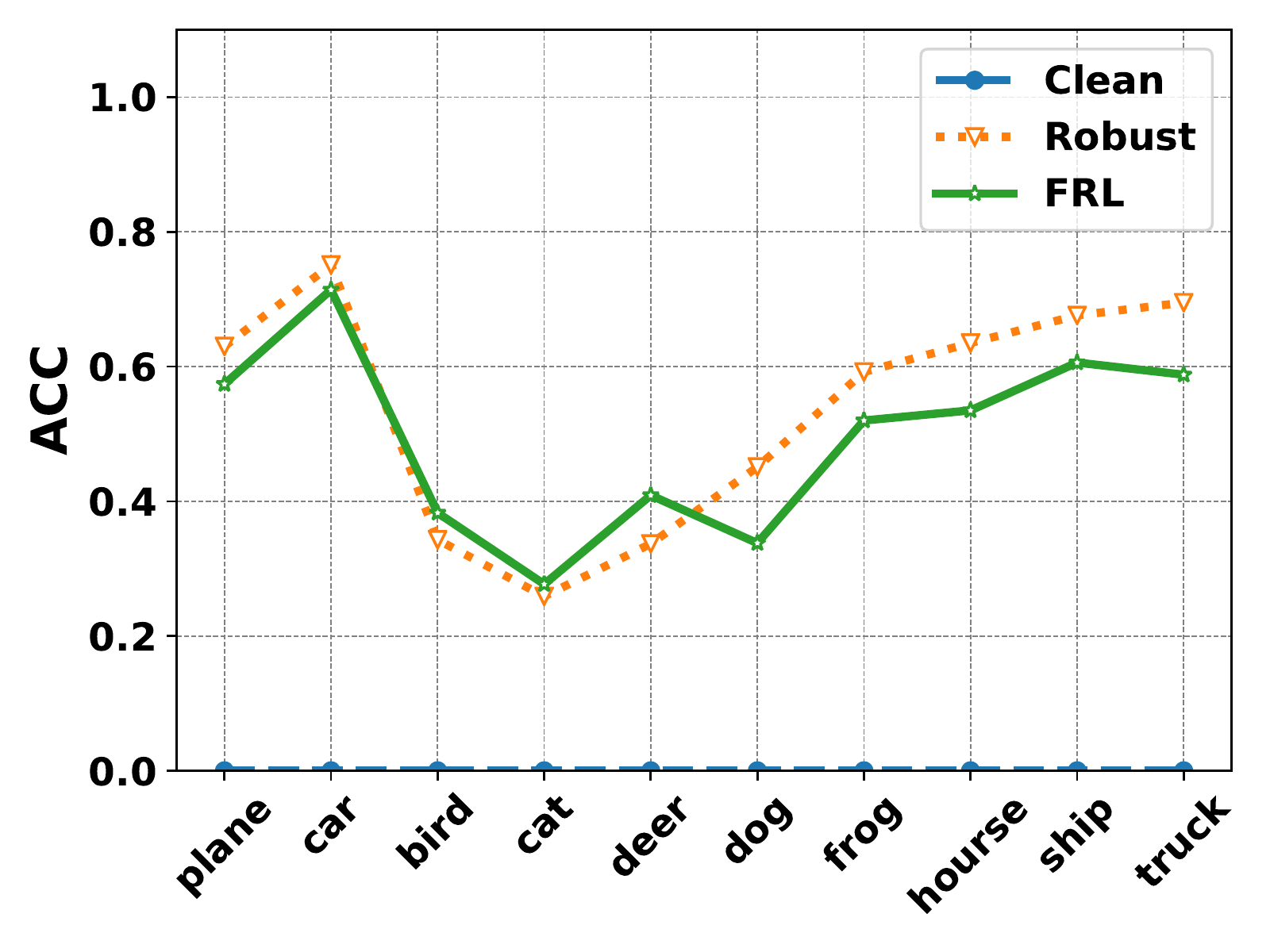}
		\caption{Robust accuracy}
		\label{fig:fairness_frl}
\end{subfigure}

\caption{AT suffers from robust fairness problem where adversarially trained models make a severe disparity in accuracy and robustness among different
classes compared to the standard training. FRL improves the previously poor performance classes, but other classes are decreased.}
\vspace{-0.15in}
\label{fig:concept-fig}
\end{figure}

%% file: texts/2_relatedwork.tex
\section{Related Work} \label{sec: related}

\subsection{Adversarial Attacks}
\label{sec:adsample}
Adversarial attacks are inputs intentionally designed to mislead deep learning models but are imperceptible to humans \cite{szegedy2014intriguing,goodfellow2015explaining}. A long line of work has been proposed to attack deep learning models \cite{goodfellow2015explaining,kurakin2016adversarial,Liu2019Perceptual,Liu2020Spatiotemporal}. In general, it can be roughly divided into white-box attacks and black-box attacks. In the white-box scenario, attackers have complete knowledge of the target model and often generate attacks using the model gradient \cite{goodfellow2015explaining,madry2018towards,carlini2017towards}; as for the black-box scenario, attackers have limited model knowledge and could often only obtain the model output \cite{IlyasICML2018,NarodytskaCVPRWORK2017,AndriushchenkoARXIV2019}. In this paper, we follow the commonly-studied setting \cite{zhang19p,DBLP:conf/icml/ZhangXH0CSK20,DBLP:conf/iclr/ZhangZ00SK21} and mainly focus on defending the more challenging white-box adversarial attacks.

\subsection{Adversarial Training}
\label{sec:adtrain}
Among the adversarial defenses \cite{DBLP:conf/cvpr/XieWMYH19,cohen2019certified,jeong2020consistency,qin2019adversarial}, adversarial training \cite{KurakinARXIV2016} that injects adversarial examples during training has been proved to be one of the most effective methods against adversarial attacks. \citeauthor{madry2018towards} formulated the adversarial training as a min-max optimization issue and utilize PGD attack \cite{madry2018towards} to solve the inner maximization for generating adversarial examples. This method makes a notable advance, and many variants of adversarial training are based on a similar min-max framework \cite{zhang19p, Wang2020Improving,wu2020adversarial}. Though promising, \citeauthor{DBLP:conf/icml/XuLLJT21} found that AT introduces severe disparity of clean and robust accuracy between different classes, which is formulated as the robustness fairness problem. 
As a preliminary study, they were motivated by \cite{buolamwini2018gender,zafar2017fairness,agarwal2018reductions} and used a re-weight and re-margin framework to finetune a robust model to improve the previously poor classes. However, they decrease the performance of other classes and make the overall accuracy (both clean and robustness) drop. In this paper, we primarily focus on better understanding and mitigating the robust fairness problem. Specifically, we discover the source-class and target-class fairness phenomena and further propose the BAT framework.

%% file: texts/3_method.tex
\section{Methodology}
\label{sec:method}

In this section, we introduce the BAT framework to mitigate the robust fairness problem. We first clarify definitions and symbols in Section~\ref{sec:notation}; we then show the source-class and target class fairness phenomena during AT in Section~\ref{sec:observation};
finally, in Section~\ref{sec:BalanceAT}, we propose novel and effective BAT methods against the robust fairness problem in AT.


\subsection{Preliminaries and Notations}
\label{sec:notation}
We use the following notations in this paper.

\textbf{Input space.} Let $\mathcal{D}$ $\subset$ $\mathbb{R}^{d}$ be the input space. 
Consider an input feature $\mathbf x_i\in\mathcal{D}$ and a label $\mathbf y_i$ is the input space $\mathcal{D}=\{(\mathbf x_i,\mathbf y_i)\}_{i=1}^{n}$.

\textbf{Deep learning model.} In this paper, we consider the image classification task. Let $f_{\theta}:{\mathbf x}\to\mathbb{R}^{K}$ represents a deep  neural network classifier parameterized by $\theta$, where ${K}$ denotes the number of the output classes, $\theta$ denotes the parameters of the model. 

\textbf{Adversarial example.} We use $\mathbf x_{adv}=\mathbf x+\mathbf{\delta}$ to denote adversarial examples, where $ ||\mathbf{\delta}||_{p} \le \epsilon$. The added perturbation $\delta$ could make DNNs misclassify the input into wrong labels, \ie, $f_{\theta}(\mathbf x+\delta) \neq f_{\theta}(\mathbf x)$.

\textbf{Adversarial training.} Given an input image ($\mathbf x_i$, $\mathbf y_i$) , a model $f_\theta$ and a loss function $\ell$, we aim to build robust DNNs through the adversarial training scheme by solving the min-max optimization problem as
\begin{equation}
    \min_\theta{\sum_{i=1}^{n}\max_{\delta}{\ell(f_{\theta}(\mathbf x_{i}+\delta,\mathbf y_{i})}}.
\end{equation}

\subsection{Source-class Fairness and Target-class Fairness}
\label{sec:observation}
In this section, we first illustrate the source-class and target-class fairness phenomena for adversarial training and then draw the relation between source$\&$target-class fairness and robust fairness. For the adversarially-trained model, we select ResNet-18 \cite{he2016deep} on CIFAR-10 \cite{krizhevsky2009learning} using PGD adversarial training \cite{madry2018towards}; we use the untargeted PGD-$\ell_{\infty}$ attack with 10 steps under $\epsilon=8/255$ budgets and the step size as $2/255$ for confusion matrices and 1000 steps under $\epsilon=8/255$ budgets and the step size as $0.4/255$ for calculating the average of attack steps. \emph{More details are in the supplementary materials.}

\input{resources/fig_3_confusion}
\input{resources/fig_2_observation}

\subsubsection{Source-class Fairness}
Source-class fairness is defined as the different difficulties of adversarial example generation from each class. There are two ways to measure it quantitatively. The first is to calculate the class-wise average number of attack steps required to cause misclassification. This is a natural way of measuring, and it reflects the distance of the decision boundary from the clean example. The second way is to fix the attack strength and count robust samples of each class, and we can use the diagonal of the confusion matrix (Figure \ref{fig:confusion}) to present the quantity. We can calculate the matrix after each attack step during the attack, and the matrix from the final attack step represents the very notion of robust fairness. We can see that the first measure integrates the second measure over the attack steps dimension. In Figure \ref{fig:observation}, we empirically observe the correlation between the average number of attack steps and class-wise robust accuracy.

\subsubsection{Target-class Fairness} 
Target-class Fairness is defined as the target class tendencies when generating adversarial examples. This quantity is calculated by the distance from the class distribution of generated adversarial examples to uniform distribution (as shown in Figure \ref{fig:observation}), and the class distribution can be calculated from the sum of the row columns in the confusion matrix (Figure \ref{fig:confusion}). In Figure \ref{fig:observation}, we see the inverse correlation between the distribution probability density and the relative class robust performance. We conjecture that this quantity is closely tied to robust fairness, which should be addressed during adversarial training.

Here we establish the correlation between source/target class fairness and robust fairness. We conjecture that addressing the source/target fairness problems in adversarial training is important to robust fairness. Therefore, we propose a new adversarial training paradigm BAT that considers source-class and target-class fairness simultaneously.


\subsection{Balance Adversarial Training}
\label{sec:BalanceAT}
In this section, we introduce our proposed Balance Adversarial Training (BAT) framework as shown in Figure \ref{fig:framwork}, where we balance both source-class and target-class fairness. 

\input{resources/fig_framwork}

\subsubsection{Balance Source-class Fairness}
We attempt to balance the number of attack steps required to break the model by adjusting the attacking strength of each class with different perturbations. Studies \cite{rade2021helper,DBLP:conf/iclr/ZhangZ00SK21} have revealed that excessive perturbations are difficult for models to fit and cause a performance drop. Therefore, we bring these samples to the decision boundaries, which would be easier and fairer for models to learn. Intuitively, for some classes that are difficult to generate adversarial examples, we should add stronger perturbations; conversely, classes that are easy to attack require fewer perturbations so that they would not be so far and “hard” to learn.

Based on the above analysis, we translate the difficulty of adversarial generation (\ie, perturbation) to the distance to decision boundaries and define two types of boundary examples. 
Given sample $\mathbf x$, let $\Phi(\mathbf x)$ denotes the maximum steps to the decision boundary, thus we have $\mathbf x^\Phi_{clean}$ as the \textbf{{last clean example}} and $\mathbf x^\Phi_{adv}$ as the \textbf{{first
adversarial example}}. Specifically, $\mathbf x^\Phi_{clean}$ denotes the ``last'' instance that can be rightly classified by models after adding perturbations, and $\mathbf x^\Phi_{adv}$ denotes the ``first'' instance that is misclassified by models after perturbing. Therefore, these two types of perturbed examples are located close to the decision boundaries, which can be referred to as \emph{boundary examples}. For each class, we adversarially perturb their samples with different strengths (perturbations) to generate the two types of boundary examples so that we could ensure that each class contains both misclassified and correctly classified samples with similar learning hardness for models. Therefore, based on the standard AT framework of TRADES  \cite{zhang19p}, we can improve the source-class fairness using $\mathcal{L}_{\mathtt{source-class}}$ as
\begin{equation}
\begin{split}
    \mathcal{L}_{\mathtt{source-class}} = \min_{\theta}\sum_{i=1}^{n}\{ {CE}(f_{\theta}(\mathbf x^\Phi_{clean,i}),\mathbf y_i)+ \\ \beta\max_{}KL(f_{\theta}(\mathbf x_i) f_{\theta}(\mathbf x^\Phi_{adv,i}))\},
    \label{eq:source-loss}
\end{split}
\end{equation}
where ${CE}$ is the cross-entropy loss, ${KL}$ is the Kullback–Leibler (KL) divergence, and $\beta$ is a balancing parameter. $\mathcal{L}_{\mathtt{source-class}}$ could balance the difficulties of adversarial example generation from source classes by avoiding generating an excess of adversarial samples for easy-to-attack classes or too few adversarial samples for hard-to-attack classes. 

\subsubsection{Balance Target-class Fairness}
 
After balancing the source class fairness, our next goal is to eliminate the biased tendencies of target classes for adversarial example generation. In other words, we need to generate adversarial examples with similar confidences or probabilities towards different classes. That is, we aim to learn a fair classifier not influenced by the tendency of adversarial targets. Formally, the generated adversarial examples should follow a uniform distribution
\begin{equation}
    \min_{{\theta}}\sum_{i=1}^{n}\left\{{KL}(\mathcal{U} f_{{\theta}}({\mathbf x_i})) + {KL}(\mathcal{U} f_{{\theta}}({\mathbf x_{adv,i}}))\right\},
    \label{eq:target-obj}
\end{equation}
where $\mathcal{U}$ is a uniform distribution of samples. Inspired by fair adversarial training \cite{zafar2017fairness}, we have the following Lemma describing that the fairness of the AT process will be influenced by different adversarial perturbations.

\noindent\textbf{\emph{Lemma 1.}}~\cite{du2021robust} \emph{The fair classifier $f$ that minimizes the cross entropy loss ${CE}(f_{{\theta}}({\mathbf x}_i+\mathbf{\delta}),\mathbf y_i)$ subject to ${RD}(\mathbb{D})\leq \tau $, where ${RD}$$(\mathbb{D})$ is the risk difference over a biased distribution $\mathbb{D}$ of $X \times \Delta \times Y$.}

In Lemma 1, $X$ denotes the input features, $\Delta$ denotes the set of different adversarial perturbations, and $Y$ denotes the label set. To make Lemma 1 reach the optimal solution (\ie, training a fair classifier $f$), we can approximate the fairness constraints ${RD}$ by using the boundary fairness as follows:
\begin{equation}
\label{boundary fairness}
    C_{\mathbb{D}}(\mathbf{\theta})=\frac{1}{n}\sum^{n}_{i=1}(\mathbf{{\delta}}_{i} -\hat{\mathbf{{\delta}}}_{i})d_{\mathbf{\theta}}({\mathbf x}_{i}),
\end{equation}
where $\mathbf{\delta}_{i}$ is the adversarial perturbation of $\mathbf x_{i}$, $\hat{\mathbf{\delta}_{i}}$ denotes the mean value of the different adversarial perturbations (different steps of attacks) added on $\mathbf x_{i}$, and $d_{\mathbf{\theta}}({\mathbf x}_i)$ indicates the distance of ${\mathbf x}_i$ to the classifier boundary of $f$.

According Lemma 1 and the boundary fairness, the overall function can be written as
\begin{equation}
     \mathcal{L}_{\mathtt{total}}=\mathcal{L}_{\mathtt{source-class}}+\alpha (\frac{1}{n}\sum_{i=1}^{n}(\mathbf{\delta}_{i}-\hat{\mathbf{\delta}}_{i})d_{\mathbf{\theta}}({\mathbf x}_{i}) - \tau )^2,
\end{equation}
where $\alpha$ is the trade off parameter. Since the distance between $\mathbf x^{\Phi}_{clean}$ and $\mathbf x^{\Phi}_{adv}$ to $\hat{\mathbf{\delta}}$ is closer than the distance between the clean example $x$ and the maximum adversarial example $\mathbf x_{adv}$ (generated by the fixed and largest perturbations), the value of the fairness loss is less. 

In this way, we notice the target-class fairness is related to the boundary samples and Eq.\eqref{eq:target-obj} can be rewritten as
\begin{equation}
\begin{split}
    \mathcal{L}_{\mathtt{target-class}} = 
    \min_{{\theta}}\sum_{i=1}^{n}\{{KL}(\mathcal{U} f_{{\theta}}({\mathbf x^\Phi_{clean,i}})+  \\KL(\mathcal{U} f_{{\theta}}({\mathbf x^\Phi_{adv,i}})\}.
    \label{eq:target-loss}
\end{split}
\end{equation}
To sum up, by uniforming the distribution of boundary examples, we could improve the target-class fairness, so that the target tendency of poor-performing classes could be reduced and the well-performing classes could be improved. 

\input{resources/algorithm}

\subsubsection{Overall Training}
Based on the above analysis, we then illustrate the overall training of our BAT framework (\emph{c.f.} Algorithm 1). In particular, we dynamically adjust the perturbation size to the boundary samples to balance Source-class fairness; we adopt the standard min-max framework and uniform distribution constraint to the boundary samples to further balance Target-class fairness. The overall training objective is shown as
\begin{equation}
\begin{split}
\mathcal{L}_{\mathtt{total}} = \mathcal{L}_{\mathtt{source-class}} + \alpha\mathcal{L}_{\mathtt{target-class}},
\end{split}
\label{eq:overall-loss}
\end{equation}
where $\alpha$ is the balancing parameter. In Algorithm 1, $\mathcal{N}(\mathbf{0}, \mathbf{I})$ generates a random unit vector of $d$ dimension, $\xi$ is a small constant. For each min-batch of data $B=\{(\mathbf x_{i},\mathbf y_{i})\}_{i=1}^{m}$, we use white-box untargeted PGD attacks to generate adversarial examples. Under the maximum PGD step K, we stop it when the samples are attacked successfully ($\arg\max_{i} f (\Tilde{ {\mathbf x_i} }) \neq \mathbf y_i$) and get $\mathbf x^\Phi_{clean}$ and $\mathbf x^\Phi_{adv}$.

%% file: resources/fig_3_confusion.tex
\begin{figure}[h]
\centering
\begin{subfigure}[h]{0.49\linewidth}
		\includegraphics[width=0.9\textwidth,height=0.73\textwidth,trim=10 10 10 10,clip]{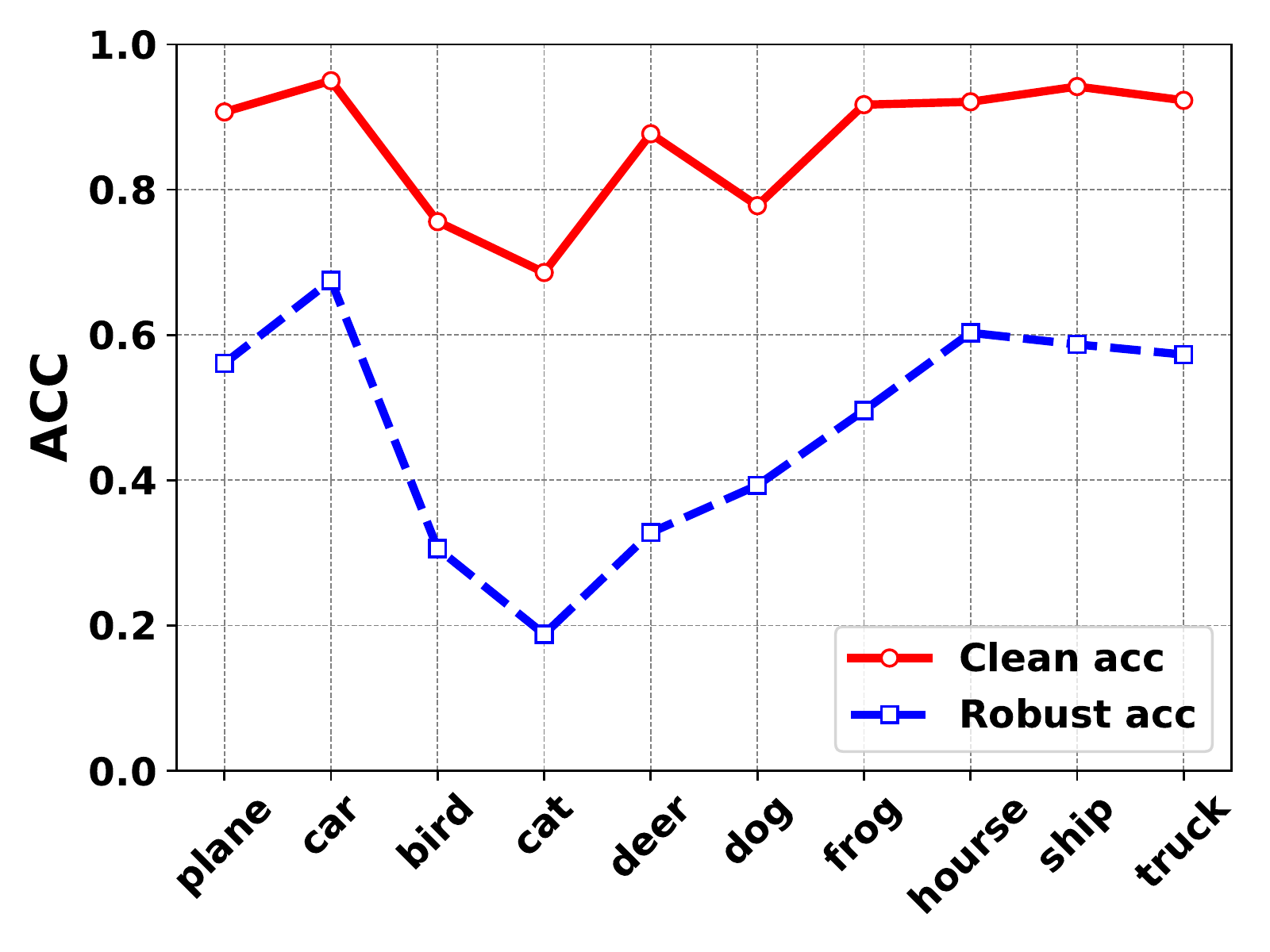}
		\caption{}
		\label{fig:Line_chart}
\end{subfigure}
\begin{subfigure}[h]{0.5\linewidth}
		\includegraphics[width=\textwidth,height=0.72\textwidth,trim=10 10 10 0,clip]{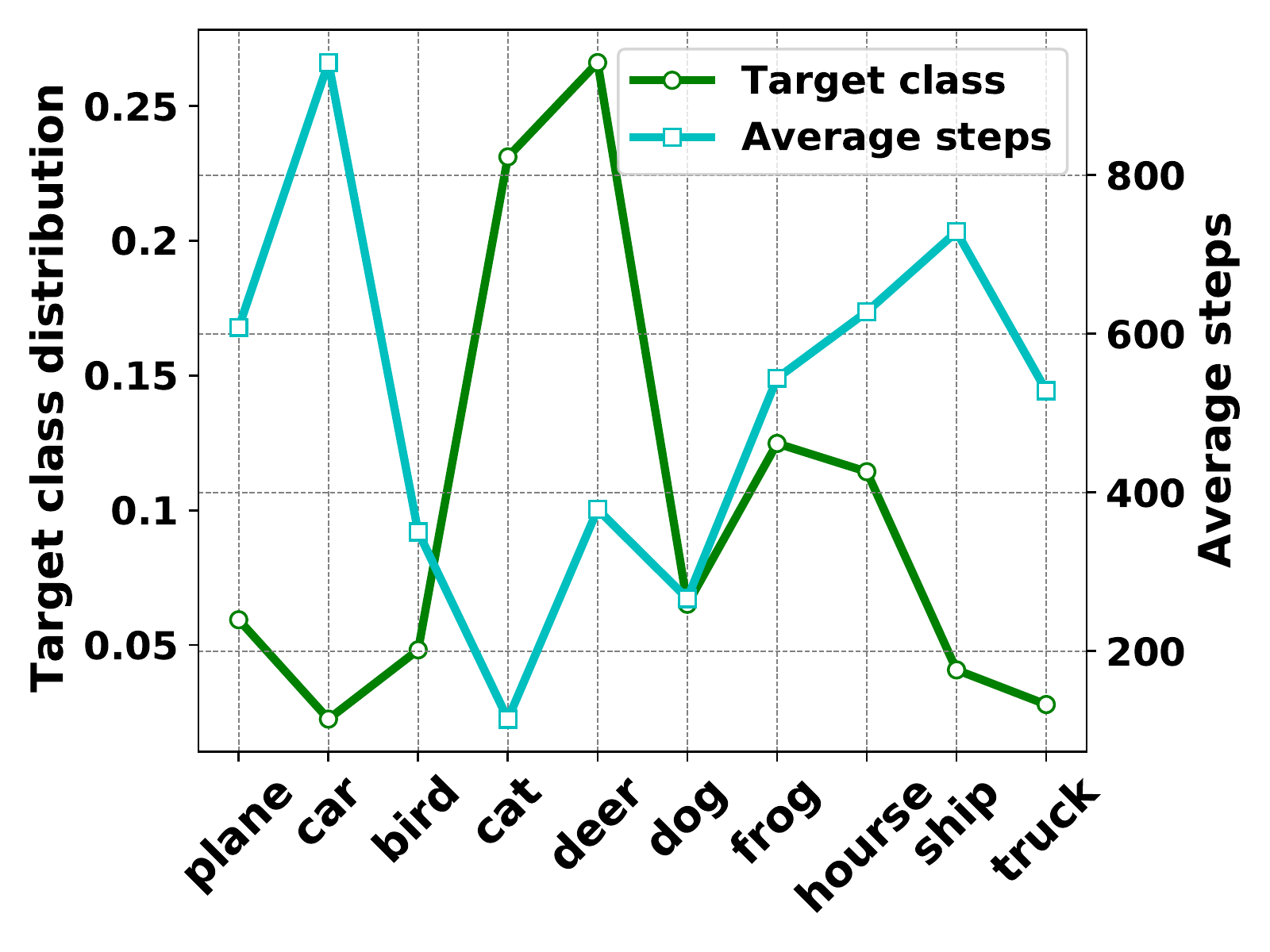}
		\caption{}
		\label{fig:Line_chart}
\end{subfigure}
\caption{Source-class fairness and Target-class fairness are both related to the robust fairness. (a): robust fairness of AT; (b): the average number of attack steps of source-class and the distribution probability of target-class.
}
\label{fig:observation}
\end{figure}

%% file: resources/fig_2_observation.tex
\begin{figure}[h]
\centering
\begin{subfigure}[t]{0.49\linewidth}
		\includegraphics[width=\textwidth,trim=10 5 30 30,clip]{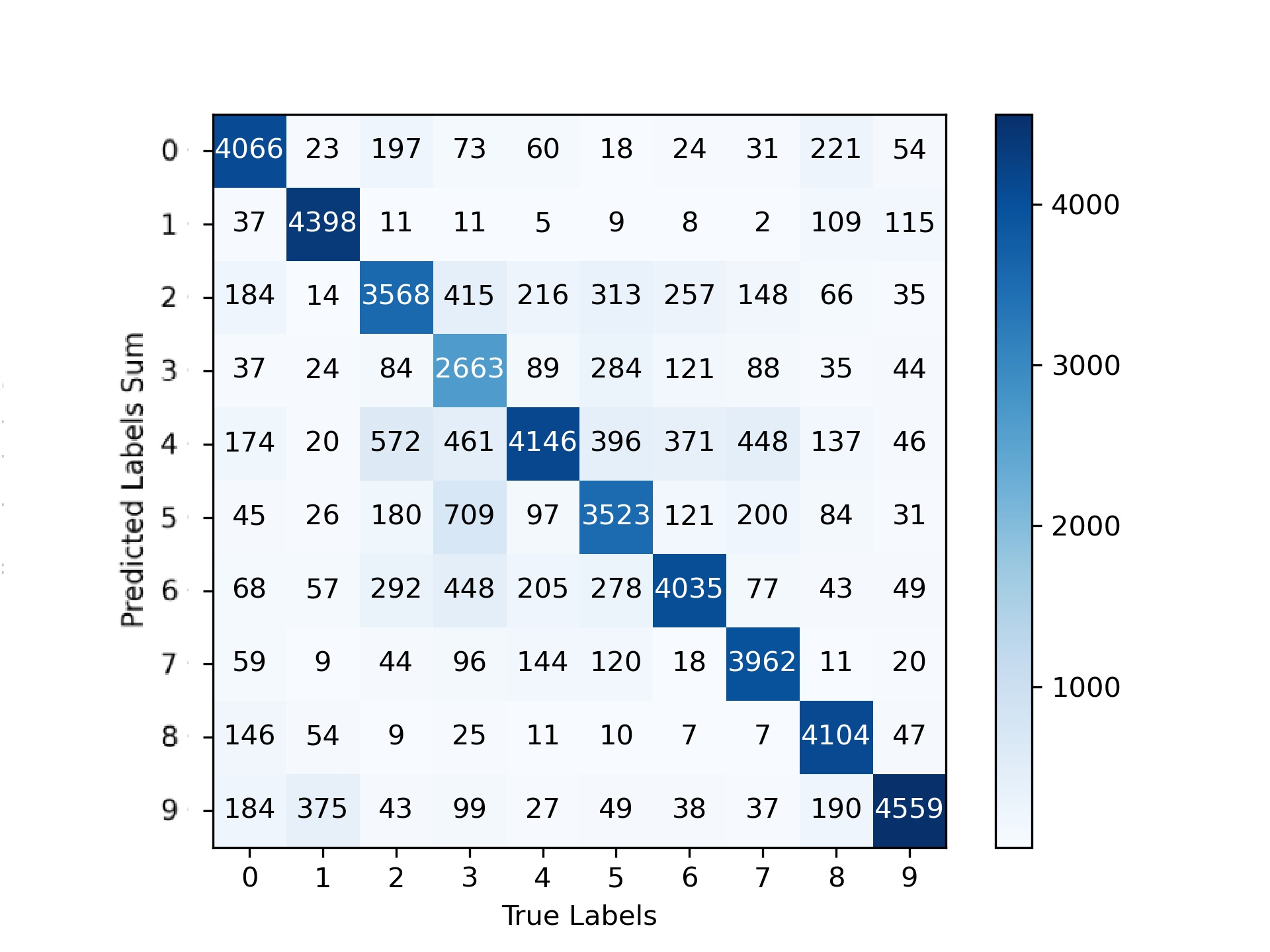}
		\caption{Epoch 50}
		\label{fig:Line_chart}
\end{subfigure}
\begin{subfigure}[t]{0.49\linewidth}
		\includegraphics[width=\textwidth,trim=10 5 30 30,clip]{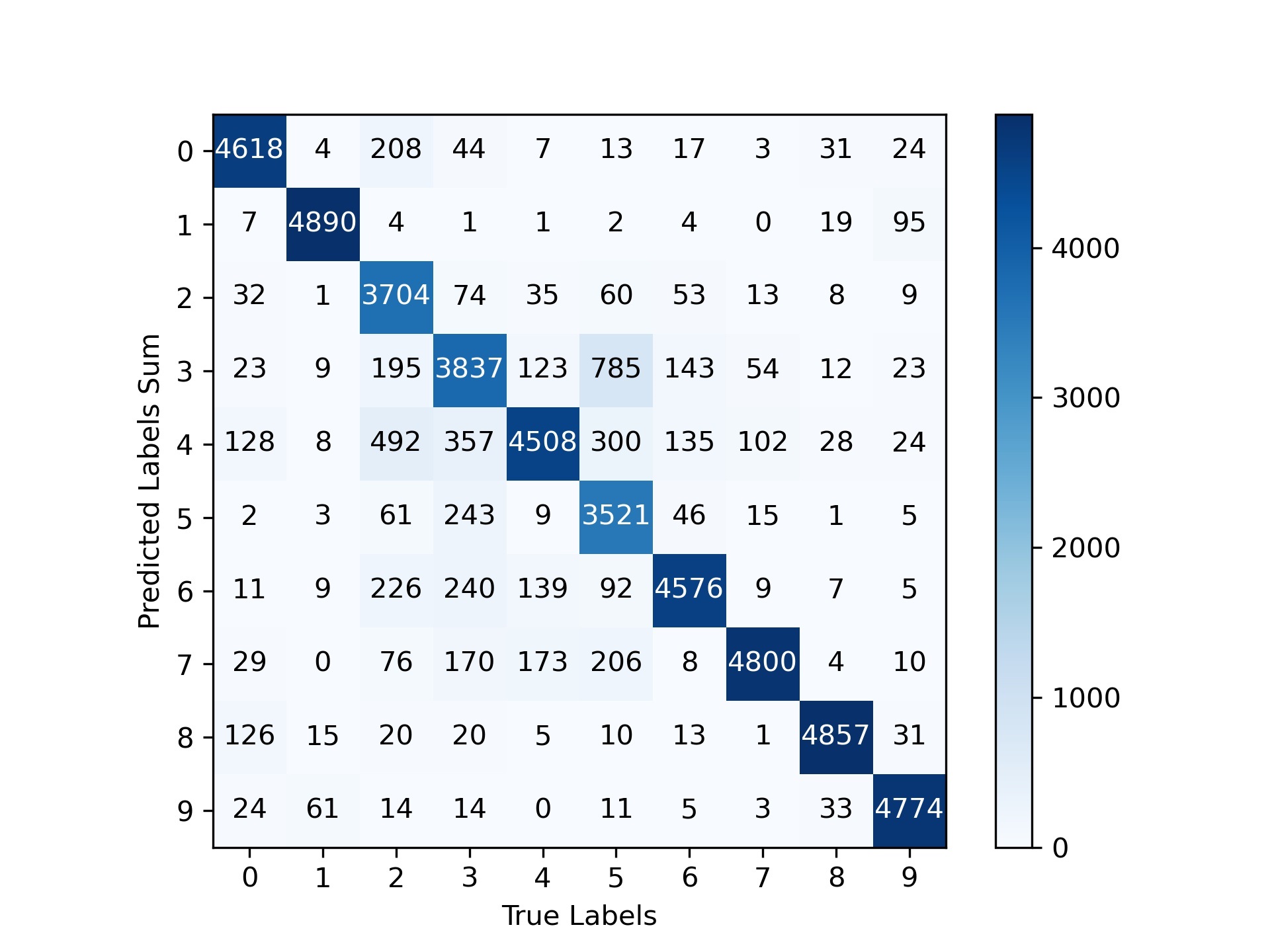}
		\caption{Epoch 60}
		\label{fig:confusion_b}
\end{subfigure}
\caption{Confusion Matrix of training data during adversarial training on different epochs.}
\vspace{-0.15in}
\label{fig:confusion}
\end{figure}

%% file: resources/fig_framwork.tex
\begin{figure*}[h]
\vspace{-0.2in}
\centering
\includegraphics[width=0.9\textwidth,trim=120 110 12 12,clip]{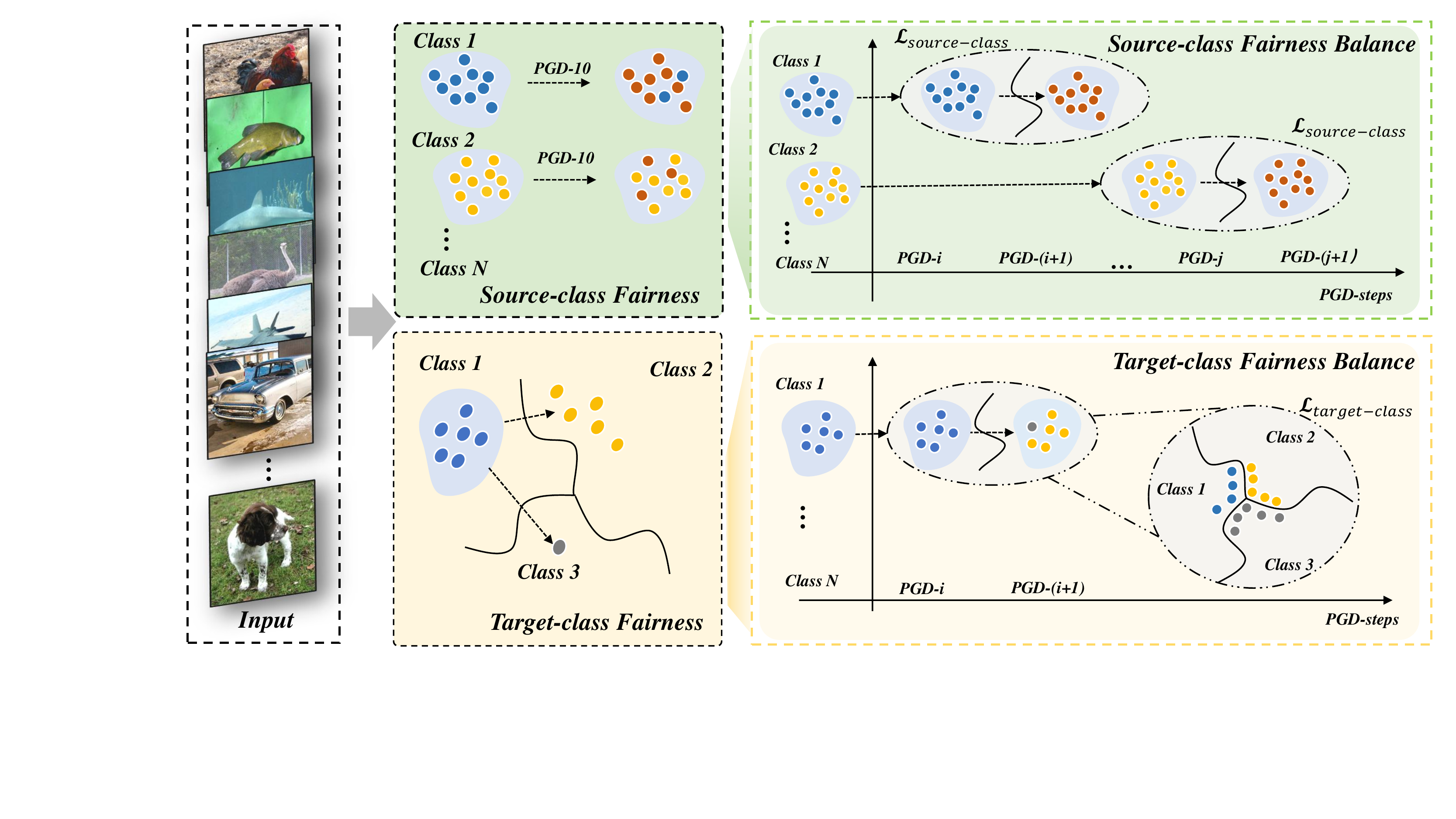}

\caption{
Framework overview. To mitigate source-class fairness, we generate adaptive perturbations where we balance the strength of adversarial attack on each class, so that we could bring these samples to the decision boundaries which would be easier and fairer for models to learn. To balance target-class fairness, we force the generated adversarial examples follow a uniform distribution, so that we could yield a fairer classifier not influenced by the tendency of adversarial targets.
}
\label{fig:framwork}
\end{figure*}

%% file: resources/algorithm.tex
\begin{algorithm}[htb]
\label{alg:overall}
\caption{Balance Adversarial Training}
\textbf{Input:} training data $(\mathbf x, \mathbf y)\sim\mathcal{D}$, batch of samples $B=\{(\mathbf x_{n},\mathbf y_{n})\}_{n=1}^{m}$, model $f_{\theta}$, loss function $\ell_{KL}$, maximum PGD step $K$, perturbation $\epsilon$, step size $\alpha$, number of epochs $T$, learning rate $\eta$ \\
\textbf{Output:} robustness network $f_{\mathbf{\theta}}$
\hrule
\begin{algorithmic}[1]
  \FOR{epoch $= 1$, $\dots$, $T$}
    \STATE Sample a mini-batch $\{(\mathbf x_i, \mathbf y_i) \}^{m}_{i=1}$ from $B$
    \STATE ${\mathbf x^\Phi_{clean,i}} \gets {{\mathbf x_i}}, {\mathbf x^\Phi_{adv,i}} \gets {{\mathbf x_i}}$
    \STATE $\Tilde{{\mathbf x_i}} \gets {\mathbf x_i} + \xi \mathcal{N}(\mathbf{0}, \mathbf{I}) $
    \WHILE{$K > 0 $}
    \IF{$\arg\max_{i} f (\Tilde{ {\mathbf x_i} }) \neq y_i$}
        \STATE \textbf{break}
    \ELSE
    \STATE ${\mathbf x^\Phi_{clean,i}} \gets \Tilde{{\mathbf x_i}}$
    \STATE $\Tilde{{\mathbf x_i}} \gets \Pi_{\mathcal{B}[{\mathbf x_i},\epsilon]}\big( \alpha(\nabla_{\Tilde{{\mathbf x_i}}} \ell_{KL}( f(\Tilde{{\mathbf x_i}}), f({\mathbf x_i})) + \Tilde{{\mathbf x_i}} \big) $ 
    \STATE ${\mathbf x^\Phi_{adv,i}} \gets \Tilde{{\mathbf x_i}}$
    \STATE $K \gets K-1$
    \ENDIF
    \ENDWHILE
    \STATE $\mathbf{\theta} \gets \mathbf{\theta} - \eta \frac{1}{m} \sum^{m}_{i - 1} \nabla_{\mathbf{\theta}} \big [ \ell_{source-class}(\mathbf x^\Phi_{clean,i}, \mathbf x^\Phi_{adv,i}) + \alpha  \ell_{target-class} (\mathbf x^\Phi_{clean,i}, \mathbf x^\Phi_{adv,i}) \big]   $
 \ENDFOR
\end{algorithmic}
\end{algorithm}
\setlength{\textfloatsep}{\baselineskip+1pt}

%% file: texts/4_experiments.tex
\section{Experiments} \label{sec: exp}

In this section, we first illustrate our experimental setups; we then compare our method with other baselines; finally, we conduct ablation studies to better understand our framework.


\subsection{Experimental Setups}
\label{sec: settings}

\textbf{Datasets and architectures.}
We choose the commonly-used datasets including CIFAR-10/100 \cite{krizhevsky2009learning} and ImageNette\footnote{We use the official data \url{https://github.com/fastai/imagenette}}. 
We use ResNet-18 \citep{he2016identity} and WRN-28-10 \cite{zagoruyko2016wide} architectures in our experiments.

\textbf{Compared baselines.}
We compare the previously proposed method FRL \cite{DBLP:conf/icml/XuLLJT21} which is the only method that address robust fairness problem to the best of our knowledge. We also consider the standard adversarial training methods, including PGD adversarial training (PGD-AT) \cite{madry2018towards} and TRADES \cite{zhang19p}. 

\textbf{Implementation details.} We use the published codes for TRADES \cite{zhang19p}\footnote{\url{https://github.com/yaodongyu/TRADES}}, FRL \cite{DBLP:conf/icml/XuLLJT21}\footnote{\url{ https://github.com/hannxu123/fair_robust}}. For FRL, we use the Reweight+Remargin under the $\tau = 0.05$ and $\tau = 0.07$ which perform best on its all settings; for the other adversarial training methods, we align our setting to the robustness benchmarks \cite{croce2020robustbench, tang2021robustart},  and set the $\epsilon=8/255$, step size $2/255$, and the maximum number of steps as 10. We keep the architecture and main hyper-parameters the same for BAT and other baselines.

\textbf{Adversarial attacks.} In this paper, we follow \cite{DBLP:conf/icml/XuLLJT21} and use PGD attacks regarding cross entropy loss with 20 steps and step size of $2/255$ to evaluate the robust fairness in our main experiment. In addition, we also adopt AutoAttack \cite{croce2020reliable} to better evaluate the robustness of our method (\emph{c.f.} supplementary material).
 
\textbf{Evaluation metrics.}
For fair comparisons, we follow \cite{DBLP:conf/icml/XuLLJT21} and use the average and worst-class error rate of standard (Avg. Std. $\&$ Worst Std.), boundary and robustness (Avg. Bndy. $\&$ Worst Bndy. and Avg. Rob. $\&$ Worst Rob.) to evaluate the robust fairness. For all these metrics, the lower the better.

\emph{We defer more details of our experimental setups to the supplementary materials.}


\input{resources/tab_1_cifar10_R18}

\subsection{Comparison with Baseline Methods}
In this section, we evaluate the robust fairness performance on ours and other baselines. Due to the space limitation, we only report the results of ResNet18 on CIFAR-10/100 and ImageNette in the main body of our paper. \emph{More results of different model architectures can be found in the supplementary materials.} Based on the results shown in Table \ref{tab:tab1}, we can draw the following \textbf{observations} and \textbf{conclusions}.

(1) For robust fairness (\ie, Worst Std., Worst Bndy., Worst Rob.), our BAT consistently outperforms other baselines by large margins on all three datasets. Compared to PGD-AT, it has around \textbf{6\%}, \textbf{10\%} and \textbf{10\%} reduction to the worst class standard error, boundary error, and robust error on CIFAR-10; for FRL, we improve the standard error, boundary error, and robust error on average \textbf{1.3\%}, \textbf{6.8\%}, \textbf{8.1\%}. More specifically, we demonstrate the class-wise performance on clean and adversarial examples in Figure \ref{fig:class}. We can observe that BAT could significantly improve the performance on \texttt{bird}, \texttt{cat} and \texttt{deer} (previously poor classes) while achieving better or similar performance on other classes. These results demonstrate the superiority of our BAT in mitigating robustness fairness problem during adversarial training.

(2) For accuracy and robustness (\ie, Avg Std., Avg. Bndy., Avg. Rob.), our BAT achieves the best performance in almost all cases. The FRL framework improves the worst class errors compared to standard AT (PGD-AT and TRADES), but it decreases the average clean and robust accuracy. For example, considering FRL (Reweight+Remargin, 0.07), the worst standard, boundary, and robust errors both decline (\ie, -3.9\%, -6.2\% and -5.0\%), but the average standard and robust errors are improved (\ie, +1.67\% and +1.64\%). Our BAT is able to avoid this drawback and leads to an overall improvement.

(3) Due to the trade-off between adversarial robustness and clean accuracy~\citep{tsipras2018robustness}, our average clean accuracy (Avg. Std.) is slightly lower than TRADES(1/$\lambda=1$), which is designed to balance the clean/robust accuracy. However, our BAT maintains a comparatively high clean accuracy with fairer robust performance. For instance, compared to TRADES under 1/$\lambda=1$ which is used to focus on better clean accuracy, our method shows slightly lower clean accuracy (0.09\%), but we achieve significantly higher robustness (\textbf{-8.42\%} on average robust errors) and fairer results on both worst class standard error, boundary error and robust error (\ie, \textbf{-1.9\%}, -\textbf{13.4\%} and \textbf{-9.4\%}); compared to TRADES under 1/$\lambda=6$, our BAT outperforms it on all metrics.

\input{resources/fig_4_each_class}

\subsection{Ablation Studies} 
\label{sec: ablation}
In this section, we provide ablation studies on our BAT. We keep the same settings with Section~\ref{sec: settings} and use CIFAR-10. 

\subsubsection{Source-class Balance of BAT.} Firstly, we study and ablate the Source-class Loss of our BAT framework. In our framework, we use the last clean sample $\mathbf x^\Phi_{clean}$ and the first adversarial sample $\mathbf x ^\Phi_{adv}$ to conduct adversarial training for better source-class fairness balancing. Here, we use ${\mathbf x}$ and ${\mathbf x_{adv}}$ instead of our source-class loss, where ${\mathbf x}$ is the clean example and ${\mathbf x_{adv}}$ is the adversarial 
example generated with fixed stable steps (\ie, 10 step numbers). Thus, the optimization objective of $\mathcal{L}_{\mathtt{source-class}}$ can be changed as follows:
\begin{equation}
    \min_{{\theta}}\sum_{i=1}^{n}\left\{{CE}(f_{{\theta}}({{\mathbf x_i}}), {\mathbf y_i})+  \beta\max_{}KL(f_{{\theta}}({\mathbf x_i})\Vert f_{{\theta}}(\mathbf x_{adv,i}))\right\}.    \label{eq:trades_form}
\end{equation}
The hyper-parameter $\beta$ and other settings are kept the same as our main experiment. From Table \ref{tab:tab3}, we can observe that our Source-class loss with decision boundary samples achieves the best performance on all evaluation metrics (No.5 vs. No.6 in Table \ref{tab:tab3}), which indicates that the decision boundary samples play a critical role in mitigating the robust fairness problem. In addition, we found that our Source-class loss has a good behavior on both standard accuracy (on average \textbf{+0.48\%} of four settings) and the robust accuracy (on average \textbf{+2.68\%} of four settings). This shows that introducing fixed perturbations for each class is harmful to the overall performance.

\input{resources/tab_3_ablation}

\subsubsection{Target-class Balance of BAT.}
Moreover, we ablate the Target-class Loss. Specifically, we first remove the target loss (No.1 in Table \ref{tab:tab3}), we then remove the uniform distribution regularization for the \emph{first adversarial samples} ${\mathbf x^\Phi_{adv}}$ (No.2 in Table \ref{tab:tab3}) and \emph{last clean sample} ${\mathbf x^\Phi_{clean}}$ (No.3 in Table \ref{tab:tab3}), respectively; finally, we use ${\mathbf x^{adv}}$ instead of our target-class loss (No.4 in Table \ref{tab:tab3}). For the samples of Target-class loss, we found that uses fixed stable steps would significantly increase the robust fairness problem, which indicates the importance of our Target-class loss. More precisely, uniformally regularizing ${\mathbf x^\Phi_{clean}}$ increases the clean accuracy while decreases the robustness; while ${\mathbf x^\Phi_{adv}}$ shows the inverse phenomenon. 
This phenomenon demonstrates that ${\mathbf x^\Phi_{clean}}$ and ${\mathbf x^\Phi_{adv}}$ are suffered from the trade-off between clean and robust accuracy. Our Target-class loss with both boundary samples has an overall improvement. Moreover, we found that only introducing Source-class loss would improve the fairness on clean data, while our Target-class loss could improve the fairness on perturbed examples. This may demonstrate that \emph{Source-class balance focuses on the performance of clean examples, and Target-class balance is more concentrated on the fairness of robustness.} We further verify this in Section \ref{sec:dataanalysis}.

%% file: resources/tab_1_cifar10_R18.tex
\begin{table*}[]
\vspace{-0.15in}
\caption{Performance of our BAT and other baselines on CIFAR-10/100 and Imagenette. Our BAT achieves the best robust fairness in almost all cases.}
\resizebox{\textwidth}{!}{
\small
\centering
\renewcommand\arraystretch{1}
\begin{tabular}{c l | cc|cc|cc}
    \toprule
    \makecell{\textbf{Dataset}} & \textbf{Method} & \textbf{Avg. Std.} & \textbf{Worst Std.} & \textbf{Avg. Bndy.} & \textbf{Worst Bndy.} & \textbf{Avg. Rob.} & \textbf{Worst Rob.} \\
    \midrule
    \multirow{6}{*}{\makecell{CIFAR-10}}  
    & PGD-AT 
    & 13.43  & 31.40  & 39.47  & 54.90  & 52.90  & 81.20  \\
    & TRADES$(1/\lambda = 1)$
    & \textbf{12.82}  & 27.80  & 38.91  & 57.80  & 51.73  & 79.70  \\
    & TRADES$(1/\lambda = 6)$
    & 15.79  & 36.30  & 31.60  & 45.60  & 47.39  & 73.80  \\
    & FRL(Reweight+Remargin, 0.05)    
    & 14.79  & 26.90  & 38.76  & 53.70  & 53.55  & 80.60  \\
    & FRL(Reweight+Remargin, 0.07)    
    & 15.10  & 27.50  & 36.16  & 48.70  & 51.26  & 76.20  \\
    & \textbf{Ours (BAT)}
    & 12.91  & \textbf{25.90}  & \textbf{31.57}  & \textbf{44.40}  & \textbf{44.48}  & \textbf{70.30}  \\
    \midrule
    \multirow{6}{*}{\makecell{CIFAR-100}}  
    & PGD-AT 
    & 40.40  & 80.00  & 36.95  & 57.00  & 77.35  & 98.00  \\
    & TRADES$(1/\lambda = 1)$
    & \textbf{40.14}  & 79.00  & 38.01  & 58.00  & 78.15  & 99.00  \\
    & TRADES$(1/\lambda = 6)$
    & 44.14  & 81.00  & 28.75  & 53.00  & 72.89  & 97.00  \\
    & FRL(Reweight+Remargin, 0.05)    
    & 43.20  & 82.00  & 29.00  & 49.00  & 73.68  & 97.00   \\
    & FRL(Reweight+Remargin, 0.07)    
    & 46.50  & 83.00  & 28.88  & 51.00  & 75.38  & 97.00  \\
    & \textbf{Ours (BAT)} 
    & 40.25  & \textbf{79.00}  & \textbf{28.68}  & \textbf{47.00}  & \textbf{70.93}  & \textbf{94.00}  \\
    \midrule
    \multirow{6}{*}{\makecell{ImageNette}}
    & PGD-AT 
    & 34.26 & 46.10 & 32.92 & 41.60 & 67.18 & 84.20 \\
    & TRADES$(1/\lambda = 1)$
    & 27.22 & 39.60 & 36.55 & 50.20 & 63.77 & 83.40  \\
    & TRADES$(1/\lambda = 6)$
    & 29.99 & 41.10 & 27.42 & 38.30 & 57.41 & 81.90  \\
    & FRL(Reweight+Remargin, 0.05)    
    & 28.05 & 40.90 & 32.00 & 42.30 & 60.05 & 81.30  \\
    & FRL(Reweight+Remargin, 0.07)    
    & 28.28 & 39.60 & 31.82 & 42.60 & 60.10 & 81.00 \\
    & \textbf{Ours (BAT)} 
    & \textbf{26.80} & \textbf{39.30} & \textbf{30.24} & \textbf{37.60} & \textbf{57.04} & \textbf{80.10}  \\
    \bottomrule
\end{tabular}
}
\label{tab:tab1}
\end{table*}

%% file: resources/fig_4_each_class.tex
\begin{figure}[htb]
\centering
\vspace{-0.1in}
\begin{subfigure}{0.23\textwidth}
\centering
\includegraphics[width=\textwidth,height=0.75\textwidth,trim=10 10 10 10,clip]{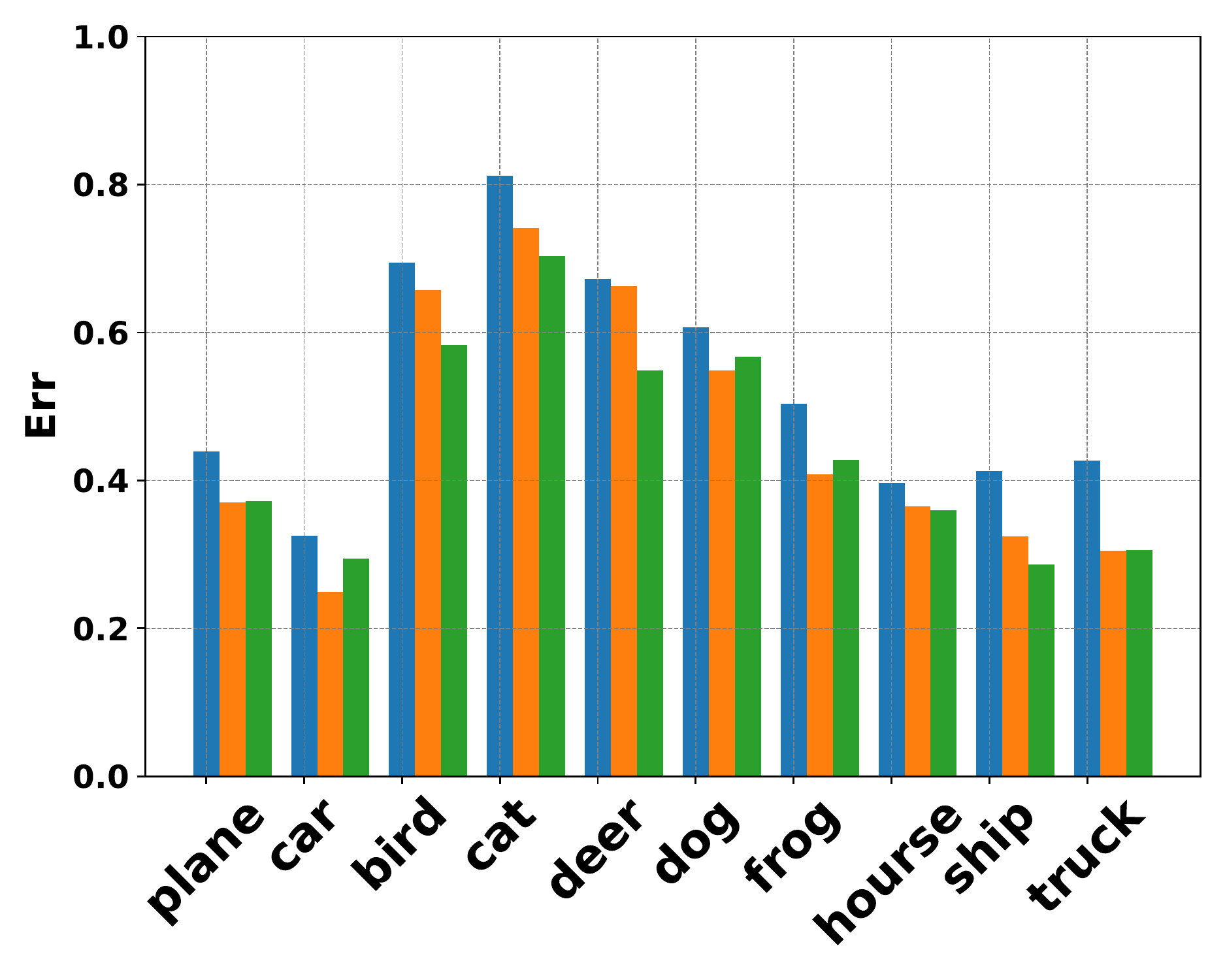}
\caption{Adversarial examples}
\end{subfigure}
\begin{subfigure}{0.23\textwidth}
\centering
\includegraphics[width=\textwidth,height=0.75\textwidth,trim=10 10 10 10,clip]{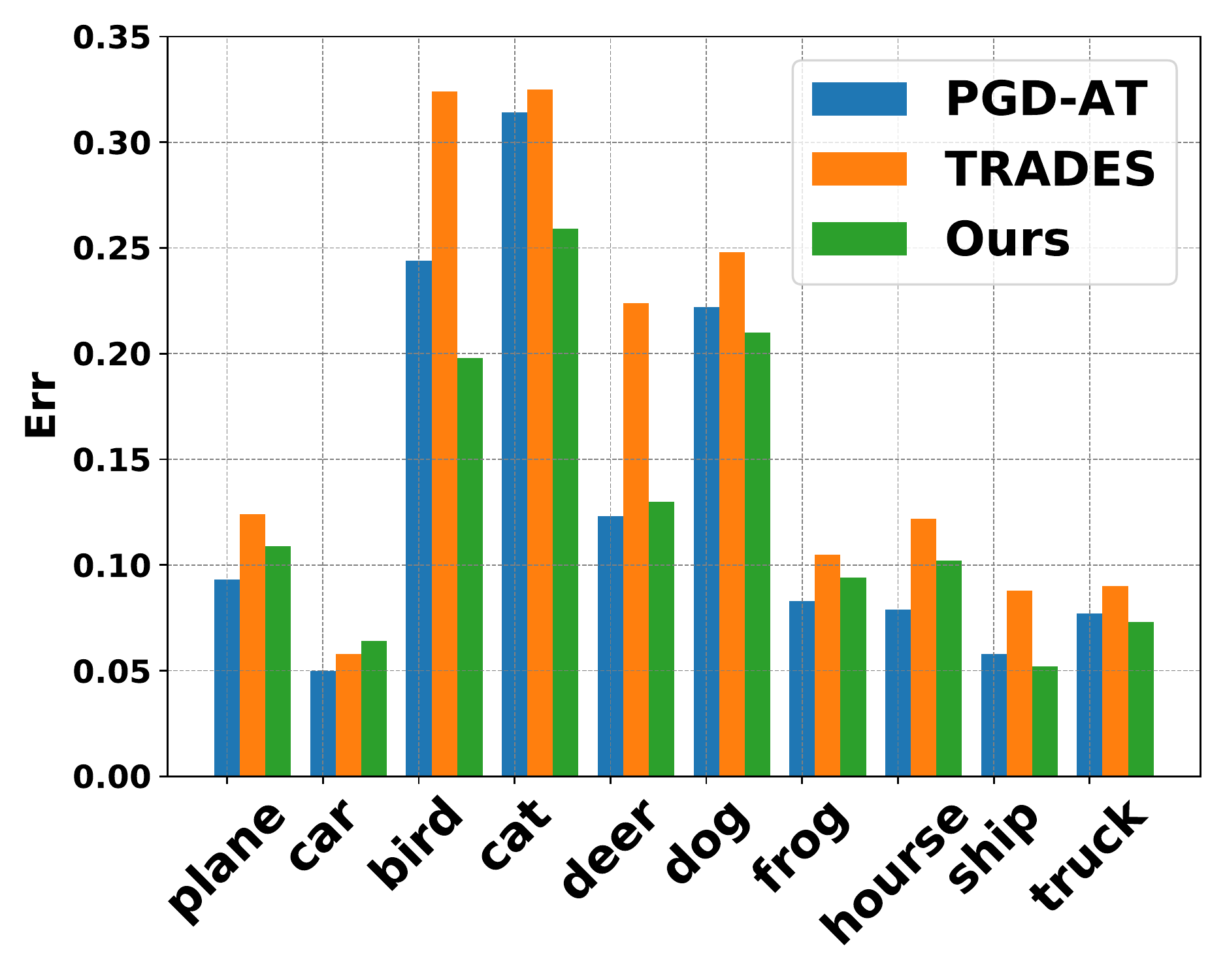}
\caption{Clean examples}
\end{subfigure}
\caption{Errors (\%) on each class of CIFAR-10 with towards clean or adversarial examples with ResNet-18 trained under PGD-AT, TRADES (1/$\lambda=6$), and BAT.} \label{fig:class}
\end{figure}

%% file: resources/tab_3_ablation.tex
\begin{table*}[t]
\vspace{-0.05in}
\caption{Average \& worst-class standard error, boundary error and robust error on Source-class Loss and Target-class Loss. No. represent the number of experiment settings.}
\resizebox{\textwidth}{!}{
\centering
\renewcommand\arraystretch{1}
\begin{tabular}{c c c| cc|cc|cc}
    \toprule
    \makecell{\textbf{Source-class Loss}} & \textbf{Target-class Loss} &\textbf{No.} & \textbf{Avg. Std.} & \textbf{Worst Std.} & \textbf{Avg. Bndy.} & \textbf{Worst Bndy.} & \textbf{Avg. Rob.} & \textbf{Worst Rob.} \\
    \midrule
    \multirow{5}{*}{\makecell{$\mathcal{L}_{\mathtt{source-class}}({\mathbf x^\Phi_{clean}}, {\mathbf x^\Phi_{adv}})$}}  
    & \emph{NA} & 1
    & 15.01  & 28.80  & 31.89  & 47.20  & 47.90  & 73.90  \\
    & $\mathcal{L}_{\mathtt{Target-class}}({\mathbf x^\Phi_{clean}})$ & 2
    & 16.74  & 30.80  & 29.34  & 43.60  & 46.08  & 74.50  \\
    & $\mathcal{L}_{\mathtt{Target-class}}({\mathbf x^\Phi_{adv}})$ & 3
    & 17.46  & 36.20  & \textbf{26.23}  & \textbf{41.30}  & \textbf{43.69}  & 70.30  \\
    & $\mathcal{L}_{\mathtt{Target-class}}({\mathbf x_{adv}})$ & 4
    & 17.51  & 37.20  & 26.59 & 42.80  & 44.10  & 72.80  \\
    & $\mathcal{L}_{\mathtt{Target-class}}({\mathbf x^\Phi_{clean}}, {\mathbf x^\Phi_{adv}})$ & \textbf{5}
    & \textbf{12.91}  & \textbf{25.90}  & 31.57  & 44.40  & 44.48  & \textbf{70.30}  \\
    \midrule
    \makecell{$\mathcal{L}_{\mathtt{source-class}}({\mathbf x}, {\mathbf x_{adv}})$}  
    & ${KL}(\mathcal{U}\Vert f_{{\theta}}({\mathbf x^\Phi_{clean}})) + {KL}(\mathcal{U}\Vert f_{{\theta}}({\mathbf x^\Phi_{adv}}))$ & 6
    & 16.65  & 40.80  & 32.11  & 47.20  & 48.76  & 77.70  \\
    \bottomrule
\end{tabular}
}
\label{tab:tab3}
\end{table*}

%% file: texts/5_analysis.tex

\section{Analysis and discussion} 
\label{sec:Analysis}
In this section, we present some analyses and discussions to better understand our BAT and the robust fairness problem.

\subsection{Is Instance-reweighting Adversarial Training helpful?} 
\label{sec:dataanalysis}
We have shown the class-reweighting scheme by FRL is not actually useful to robust fairness, but how instance-reweighting in adversarial training affect robust fairness?

\input{resources/tab_2_vsfat}

There exist several studies that exploit the instance-reweighting technique to better balance the clean/robustness trade-off of AT, and here we examine Friendly Adversarial Training (FAT) \cite{DBLP:conf/icml/ZhangXH0CSK20} and Geometry-aware Instance-Reweighted Adversarial Training (GAIRAT) \cite{DBLP:conf/iclr/ZhangZ00SK21}. FAT uses friendly adversarial samples, which are the least misclassified generated by attacks, while GAIRAT up-weights the boundary instance during AT. From Table \ref{tab:tab2}, our BAT achieves better performance on robust fairness than FAT and GAIRAT (\textbf{1.9\%}, \textbf{13.3\%}, \textbf{12.9\%} and \textbf{9.0\%}, \textbf{2.5\%}, \textbf{9.0\%}) in terms of Worst Std., Worst Bndy., and Worst Rob. Meanwhile, BAT also shows better clean and robustness trade-off performance than these two methods.

We provide a closer inspection of these results. The key observation is that FAT and GAIRAT re-weighting directions are different: the former decreases the loss by boundary examples, while the latter increases it. From Table \ref{tab:tab2}, we can see the trade-off between worst clean and worst robustness. If we refer to Table \ref{tab:tab1}, we can see the performance of FAT is close to TRADES 1/$\lambda=1$, which has weak regularization of AT, and that of GAIRAT is close to PGD-AT, which adds more perturbation to adversarial examples. We can conclude that excessive perturbations would cause a performance drop in clean accuracy. Thus, the source-class fairness term in BAT may play a similar role to the instance reweighting scheme of FAT, and we see the improvement in terms of clean fairness (Section \ref{sec: ablation}). This also verifies our hypothesis on source-class fairness, which is more closely related to clean accuracy. Thus, FAT and GAIRAT could only improve robust fairness to some extent due to the ignorance of target-class fairness. Despite that target fairness cannot be used alone to get robustness, combining the terms are important to the robustness measure in fairness, which makes BAT better than all baselines and other instance-reweighting methods.


\subsection{Does the Devil Exist in the Dataset?}

\input{resources/tab_4_cifar6}
\input{resources/fig_5_badcase}

We notice that the robust fairness problem relates to the inherent difficulty in robust learning. Since AT models often show weak performance on specific classes within a dataset, we try to remove these classes from the dataset and then re-train models. In particular, we adversarially train ResNet-18 models using PGD-AT on CIFAR-10, where we erase previously poor classes (\ie, 2, 3, 4, and 5) or previously good classes (\ie, 1, 6, 8, and 9), respectively. As shown in Table \ref{tab:tab4}, models trained on datasets without previously poor classes show an obvious decrease in worst class metrics, which indicates that the robust fairness problem is somewhat mitigated by removing the ``hard'' classes. 

Some of the improvement in fairness comes from the mitigation of spurious correlation \cite{sagawa2020investigation}. We visualize some bad cases of these hard classes (\eg, 2, 3, 4, and 5 in CIFAR-10) in Figure \ref{fig:badcase}, where models trained by our BAT could correctly recognize these images while PGD-AT fails. We see PGD-AT suffers from the spurious correlation with the background. For example, the test image \texttt{Bird} has a similar blue background and two wings to the class \texttt{Plane}. 

%% file: resources/tab_2_vsfat.tex
\begin{table}[htb]
\vspace{-0.15in}
\centering
\caption{Comparison with FAT and GAIRAT on CIFAR10. Our BAT shows better performance on all metrhics, indicating the importance of addressing both source-class and target-class fairness.}
\resizebox{0.45\textwidth}{!}
{
\renewcommand\arraystretch{1}
\begin{tabular}{l |c c |c c |c c} 
\toprule
& \textbf{Avg. Std.} & \textbf{Worst Std.} & \textbf{Avg. Bndy.} & \textbf{Worst Bndy.} & \textbf{Avg. Rob.} & \textbf{Worst Rob.}  \\  
\midrule
FAT & \textbf{12.54}  & 27.80  & 38.98  & 57.70  & 51.52  & 83.20  \\ 
GAIRAT & 16.74  & 34.90  & 32.42  & 46.90  & 49.16  & 79.30 \\ 
\textbf{Ours} & 12.91  & \textbf{25.90}  & \textbf{31.57}  & \textbf{44.40}  & \textbf{44.48}  & \textbf{70.30} \\ 
\bottomrule
\end{tabular}}
\label{tab:tab2}
\end{table}

%% file: resources/tab_4_cifar6.tex
\begin{table}[h]
\vspace{-0.05in}
\centering
\caption{Training with all dataset, training without previously poor classes (\ie, 2, 3, 4, and 5), and training without previously good classes (\ie, 1, 6, 8, and 9).}
\resizebox{0.45\textwidth}{!}
{
\renewcommand\arraystretch{1}
\begin{tabular}{l |c c |c c |c c} 
\toprule
& \textbf{Avg. Std.} & \textbf{Worst Std.} & \textbf{Avg. Bndy.} & \textbf{Worst Bndy.} & \textbf{Avg. Rob.} & \textbf{Worst Rob.}  \\  
\midrule
All classes & 13.43 & 31.47 & 39.47 & 54.91 & 52.90 & 81.20 \\
Without 2,3,4,5 & 5.15 & 6.43 & 30.52 & 38.23 & 35.67 & 49.28  \\ 
Without 1,6,8,9 & 17.83 & 32.64 & 40.09 & 52.59 & 57.92 & 80.52 \\ 
\bottomrule
\end{tabular}}
\label{tab:tab4}
\end{table}

%% file: resources/fig_5_badcase.tex
\begin{figure}[h]
\vspace{-0.15in}
\centering
\includegraphics[width=0.88\linewidth]{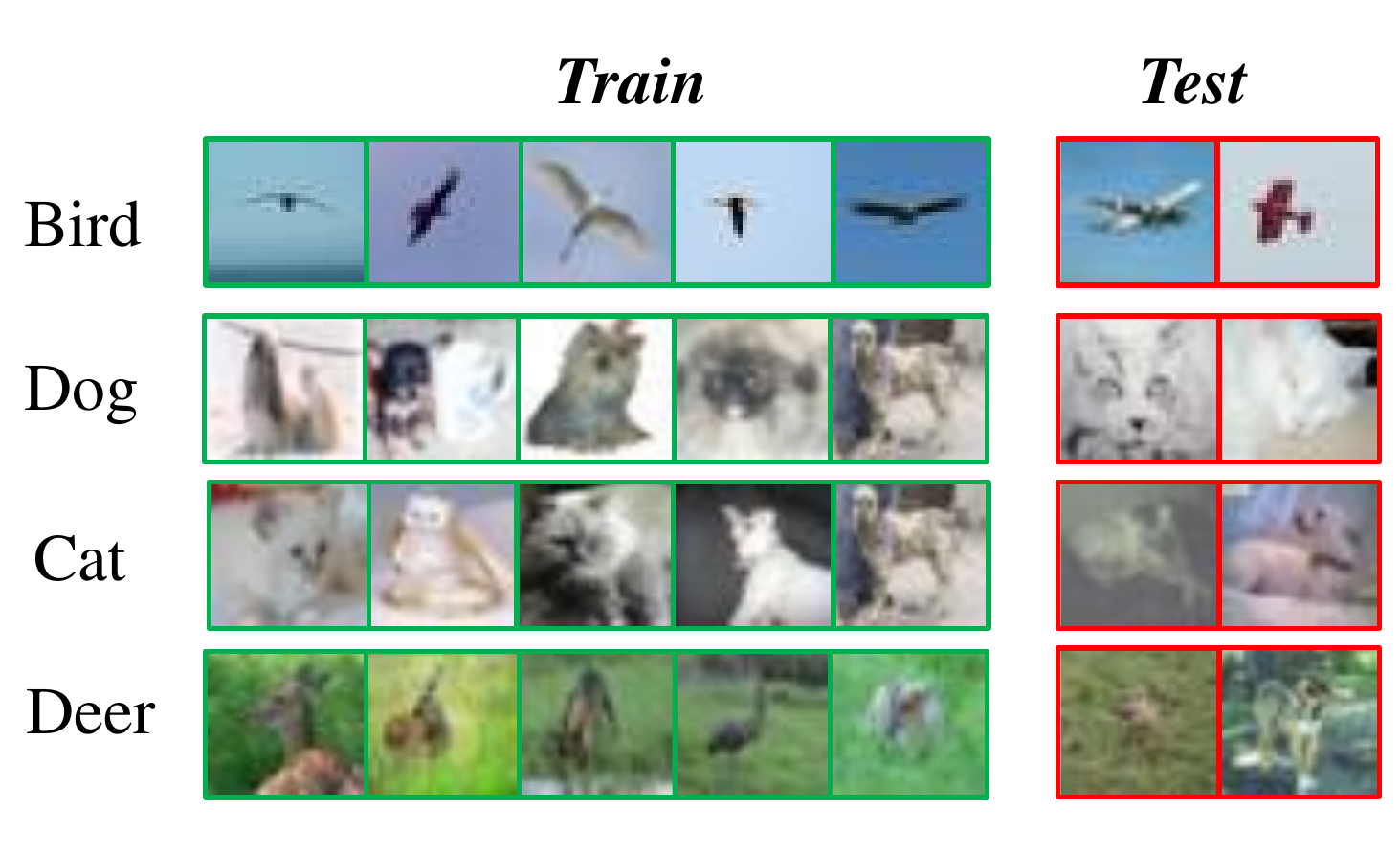}

\caption{Bad case study. Test images that are failed by PGD-AT but correctly classified by our BAT are quite similar to the
training examples of wrongly classified classes.}
\label{fig:badcase}
\end{figure}

%% file: texts/5_conclusion.tex
\section{Conclusion} \label{sec: discussion}

We find the correlation of robust fairness with source-class and target-class fairness. Based on the observation, we further propose Balance Adversarial Training to mitigate the robust fairness in adversarial training, where we simultaneously balance source-class and target-class fairness. Experiments demonstrate that BAT significantly improves robust fairness. In the future, we will study the generalization of robust fairness under more attacks and design metrics considering clean, robustness, and robust fairness.